# Active Learning for Matching Problems

**Laurent Charlin** LCHARLIN@CS.TORONTO.EDU
**Richard Zemel** ZEMEL@CS.TORONTO.EDU
**Craig Boutilier** CEBLY@CS.TORONTO.EDU
Department of Computer Science, University of Toronto, Toronto, ON M5S 3H5

## Abstract

Effective learning of user preferences is critical to easing user burden in various types of matching problems. Equally important is active query selection to further reduce the amount of preference information users must provide. We address the problem of active learning of user preferences for matching problems, introducing a novel method for determining probabilistic matchings, and developing several new active learning strategies that are sensitive to the specific matching objective. Experiments with real-world data sets spanning diverse domains demonstrate that matching-sensitive active learning outperforms standard techniques.

## 1. Introduction

The burgeoning interest in recommender systems has led to a plethora of techniques for predicting *user preferences* or ratings for unseen *items* (e.g., products in e-commerce applications). *Collaborative filtering (CF)* methods (Goldberg et al., 1992) have proven especially popular and have attained impressive performance (Koren, 2009). In practice, however, recommendations must not only account for user preferences in isolation; one usually has to tradeoff preferences for recommended items with various *constraints* or *objectives*. For example, an online retailer may want to limit the number of recommendations (to different users) for any particular item so stock is not depleted (which would create unsatisfied customers). The same retailer may wish to facilitate serendipitous purchases by ensuring items recommended to any single user are diverse (McNee et al., 2006).

In this paper we focus on *match-constrained recommendation*, where the quality of a *set of recommendations* or *matching* is measured relative to constraints or objectives that account for the *entire set of users* to whom an item is recommended, the *entire set of items* recommended to a single user, or both. These are traded off against the predicted degree of preference of individual recommendations. Match-constrained recommendation has wide application. For example, consider the problem of assigning papers to reviewers: given preferences (self-assessed expertise) of reviewers for certain papers, we want to find the best assignment of papers to reviewers. Recent work has used learning techniques such as CF to predict missing preferences—allowing reviewers to specify preferences for only a small selection of papers—and finding high quality matches subject to specific "collective" constraints on the matching (e.g., number of papers per reviewer, number of reviewers per paper) (Conry et al., 2009; Charlin et al., 2011).

Eliciting preferences (e.g., in the form of item ratings) imposes significant time and cognitive costs on users. In domains such as paper matching, product recommendation, or online dating, users will have limited patience for specifying preferences. While learning techniques can be used to limit the amount of required information in match-constrained recommendation, the intelligent selection of preference queries is just as important in reducing user burden. It is this problem we address in this paper. We frame the problem as one of *active learning*: our aim is to determine those preference queries with the greatest potential to improve the quality of the matching. This is a departure from most work in active learning, and, specifically, approaches tailored to recommender systems (as we discuss below) where queries are selected to improve the overall quality of ratings prediction. We develop techniques that focus on queries whose responses will directly impact—possibly indirectly by changing predictions—the matching objective itself.

In this paper we propose several new active learning methods for match-constrained recommendation. In contrast to previous active approaches, our methods are sensitive to the matching objective. We also propose a new probabilistic matching technique that accounts for uncertainty in predicted preferences when constructing a matching. Finally, we test our methods on several real-life datasets for online dating, conference reviewing, and assigning jokes to users. Our results show that active learning methods that are sensitive to the matching task significantly outperform a standard active learning method. Furthermore, we show that our probabilistic methods can be successfully leveraged by active learning.

## 2. Related Work

We define our problem formally below, but informally, assume a set of *users* must be matched to *items* subject to certain constraints or objectives on the overall matching. *Items* are construed broadly to refer to anything to which users might be matched: products in e-commerce recommender systems, papers in reviewer matching problems, even other users in roommate assignment, stable marriage or online dating. User *preferences* over items are represented by *suitability scores* which reflect the quality of matching a given item to a user "in isolation." Again, suitabilities should be interpreted broadly as user preference, expertise, or some other application-specific metric.

Matching problems have been studied in economics for decades, where the focus has been on *incentives* and *stability*, especially in two-sided matching domains where both users and items (e.g., other users) express preferences over the other. Classic examples include stable marriages and college admissions (Gale & Shapley, 1962), medical resident to hospital matching (Roth, 1984) and (one-sided) housing markets (Hylland & Zeckhauser, 1979). This work typically assumes all preferences have been specified.

Considerable work in information retrieval and machine learning has dealt with predicting suitabilities given only a partial set of scores or other relevant data. CF is, of course, a prime example (Goldberg et al., 1992). Recent work on reviewer-paper matching—a domain we consider here—includes work using coauthorship graphs (Rodriguez & Bollen, 2008), language and topic models (Mimno & McCallum, 2007), and CF (Conry et al., 2009; Charlin et al., 2011). The latter work in particular shows that high-quality matchings can be constructed while eliciting only a small set of suitabilities, and that tuning the learning objective to account for the matching objective can improve match quality with limited data.

Active learning is a rich field (e.g., see (Settles, 2009)

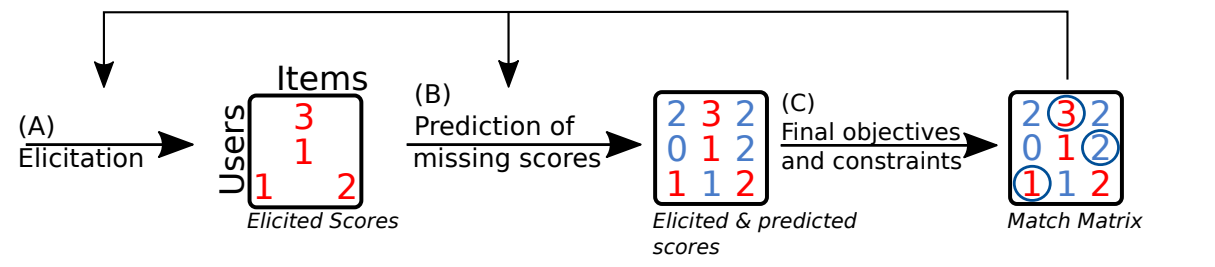

*Figure 1.* Match-constrained recommendation framework. Observed scores are shown in red, unobserved in blue, and assigned are circled.

for a recent survey). Closest to our work are active learning methods developed for CF (Boutilier et al., 2003; Jin & Si, 2004; Harpale & Yang, 2008). These methods consider objectives that deal with recommendation sets (somewhat akin to our matching objective). Rigaux (2004) considers an iterative elicitation method for paper matching using neighborhood CF, but requires an initial partitioning of reviewers, and elicits scores (with the aim only of improving score prediction quality) for the same papers from all reviewers in a partition. In our work, we need not partition users, and we focus on matching quality rather than prediction accuracy. Our approach is thus conceptually similar to CF methods trained for specific recommendation tasks (e.g., (Weimer et al., 2007)). Finally, Bayesian optimization has been used for active learning recently (Brochu et al., 2009); however, these methods assume a continuous query space and some similarity metric over item space, hence are not readily adaptable to our match-constrained problems.

## 3. Match-constrained Recommendation

In this section, we describe a general framework for exploiting learning in match-constrained recommendation, adopting the model and principles used in recent work on paper matching (Conry et al., 2009; Charlin et al., 2011). We then introduce a new model for probabilistic matching within this framework that we exploit when considering active learning in Sec. 4.

### 3.1. Matching Framework

Our aim is to determine high quality matchings of items to users while requiring users to specify preferences or *suitabilities* over only a small fraction of all items. Our overall framework is illustrated in Fig. 1, and comprises three steps, which may be iterated in the active learning setting (Sec. 4). We first elicit suitability scores from users for certain items (A); we then learn some model (e.g., CF) to predict unobserved suitabilities (B); finally, some form of optimization is used to determine a matching meeting certain objectives and constraints, using both observed and predicted suitabilities (C). We first formalize the model, then discuss components (B) and (C) in more detail.

**Model Formulation** We formalize our problem as follows. Let $r \in \mathcal{R}$ refer to users (e.g., reviewers, consumers, etc.), $p \in \mathcal{P}$ to items (e.g., papers, products, other users), with $|\mathcal{R}| = N$ and $|\mathcal{P}| = M$. Every user-item pair has a *suitability score* $s_{rp}$, the set of which forms a *suitability matrix* $S \in \mathbb{R}^{N \times M}$. Only a subset of the suitabilities are observed, namely, those elicited from users. We denote this by $S^o$, and denote the observed scores for a particular user $r$ and item $p$ by $S^o_r$ and $S^o_p$, respectively. $S^u$, $S^u_r$, $S^u_p$ are the analogous collections of unobserved scores.

**Estimating Unobserved Scores** Estimating unobserved scores given a (small) set of observed scores—together with any side-information, e.g., word vectors in submitted papers—is a straightforward supervised learning task to which a variety of approaches can be applied. Conry et al. (2009), for instance, use CF to predict unknown scores in a paper matching domain. While our framework is quite general, our experiments, including those in Sec. 5 on paper matching and several other domains, suggest that *Bayesian probabilistic matrix factorization (BPMF)* (Salakhutdinov & Mnih, 2008a) is a very competitive CF method for match-constrained recommendation. So we describe it here as one method, of many, that can be used.

Standard PMF (Salakhutdinov & Mnih, 2008b) factorizes the observed score matrix, $S^o$, into two low-rank matrices, $U \in \mathbb{R}^{N \times k}$ and $V \in \mathbb{R}^{M \times k}$, where $k \ll \min(M, N)$. Unobserved scores are predicted using their product: $s^u_{ij}$ is predicted to be $u_i v'_j$. While PMF can be given a probabilistic interpretation by assuming Gaussian noise with fixed standard deviation $\sigma$, BPMF provides a Bayesian extension of PMF by assuming priors on $U$ and $V$. Apart from outperforming PMF on standard CF tasks, BPMF also provides a *distribution* over unobserved suitabilities, $\Pr(S^u|S^o, \theta)$, instead of a simple point prediction.

**Matching Optimization** Matching uses both observed and predicted suitability scores to assign items to users. We focus primarily on domains where constrained many-to-many matching is required, i.e., where an item is matched to multiple users and multiple users are matched to a single item. Paper matching is a prime example, where each paper needs a specific number of reviewers, and each reviewer must be assigned a number of papers within some range.

While many objectives and constraints can be accommodated in our framework, we illustrate it using a simple paper matching problem. Treating self-reported interest or expertise as observed suitabilities, standard learning methods like BPMF are used to predict unobserved scores. We then formulate the matching optimization (treating predicted scores as if observed) as an integer program (IP) (Taylor, 2008):

$$\text{maximize} \quad J(Y, S) = \sum_r \sum_p s_{rp} y_{rp} \tag{1}$$

$$\text{subject to} \quad y_{rp} \in \{0, 1\}, \quad \forall r, p \tag{2}$$

$$\sum_r y_{rp} \geq R_{\min}, \ \sum_r y_{rp} \leq R_{\max} \quad \forall p \tag{3}$$

$$\sum_p y_{rp} \geq P_{\min}, \ \sum_p y_{rp} \leq P_{\max}, \ \forall r. \tag{4}$$

Here the binary variable $y_{rp} \in Y$ encodes the matching of item $p$ to user $r$; a *matching* is a complete instantiation of these variables. $R_{\min}$ ($R_{\max}$) is the minimum (resp. maximum) number of users per item, while $P_{\min}$ ($P_{\max}$) represent the minimum (resp. maximum) user capacities. Of course, many other criteria can be incorporated into the matching optimization. While IPs of this form can quickly become intractable, the total unimodularity of the constraint matrix (Eqs. 2–4) allows one to use the linear programming (LP) relaxation while retaining optimality.

### 3.2. Probabilistic Matching

While the LP optimization is straightforward, and provides optimal solutions when all scores are observed, it has potential drawbacks when used with predicted scores, and specifically, when used in conjunction with active learning. First, the LP does not consider potentially useful information contained in the uncertainty of the (predicted) suitabilities. Second, it does not express the *range of possible matches* that might optimize total suitability (given the constraints).

While optimal matching given *true* scores can be viewed as a deterministic process, score prediction is inherently stochastic; and we can exploit this if our prediction model outputs a distribution over unobserved scores $S^u$ rather than a point estimate. Given inputs consisting of observed scores $S^o$ and possibly additional side-information $X$, we can express our uncertainty over the optimal matching as:

$$\Pr(Y|S^o, X, \theta) = \int Y(S^u, S^o) \Pr(S^u|S^o, X, \theta)\, dS^u, \tag{5}$$

where $\Pr(S^u|S^o, X, \theta)$ is our score prediction model (assuming model parameters $\theta$), and $Y(\cdot)$ (see Eq. 1) is the optimal matching given a fixed set of scores.

With this in hand, we overcome the limitations of pure LP-based optimization by developing a sampling method for determining "soft" or *probabilistic matchings* that reflect the range of optimal matchings given uncertainty in predicted suitabilities. While Eq. 5 expresses the induced distribution over optimal matchings, the integral is intractable as it requires solving a large number of matching problems (e.g., LPs). Instead we take a sampling approach: we independently sample each score from the posterior $\Pr(S^u|S^o, X, \theta)$ to build a complete score matrix, then solve the matching optimization (LP) using this sampled matrix. Repeating this process $T$ times provides an estimated distribution over optimal matchings. We can then average the resulting match matrices, obtaining $\overline{Y} = \frac{1}{T}\sum_{t=1}^{T} Y^{(t)}$. Each entry $\overline{Y}_{rp}$ is the (estimated) probability that user-item pair $rp$ is matched; and the probability of this match depends, as desired, on the distribution $\Pr(s_{rp}|S^o, X, \theta)$.

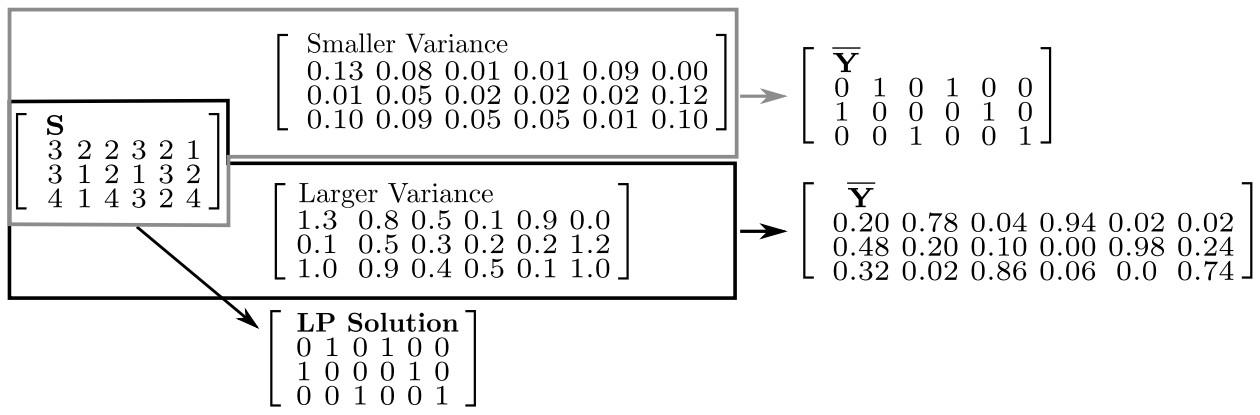

*Figure 2.* A "toy" example with a synthetic score matrix $S$, and matching results using LP (bottom) and the $\overline{Y}$ approximation with two different variance matrices.

Fig. 2 illustrates $\overline{Y}$, comparing it to the LP solution, on a randomly-generated "toy" problem with 3 reviewers and 6 papers (1 reviewer per paper, 2 papers per reviewer). Assuming a fixed predicted score matrix $S$, two versions of $\overline{Y}$ are shown, one when all estimated variances are low, the other when they are higher.[1] Note that the $\overline{Y}$ matrices respect the matching constraints by design (for visualization purposes we round matching probabilities). With low variances, $\overline{Y}$ agrees with the LP, but with higher variances, we observe the inherent uncertainty in the optimal matching; e.g., column one shows all three match probabilities to be reasonably high. In addition, the last column shows that even though the second and third users have scores that differ by 2, the high variance in their scores gives both users a reasonable probability of being matched.

## 4. Active Querying for Matching

Since it is impractical to elicit or otherwise observe the preferences of all users for all items, supervised learning, as discussed above, can be used to effectively estimate unobserved suitabilities for match-constrained recommendation (Conry et al., 2009; Charlin et al., 2011). However, little work has considered strategies for actively querying the "most informative" preferences from users, thus further reducing the elicitation burden on users. Random selection of user-item pairs for assessment will generally be sub-optimal, since query selection is uninformed by the learned model, the objective function, or any previous data. By contrast, an *active approach*, in which queries are tailored to both the current preference model *and* the current best matching, will typically give rise to better matchings with fewer queries.[2]

In this section we describe several distinct strategies for query selection: we review a standard active learning technique and introduce several novel methods that are sensitive to the matching objective. Our methods can be broadly categorized based on two properties: whether they evaluate in *score space* $S$ or *matching space* $Y$; and whether they select queries with the *maximal value* M, or *maximal entropy* E.

*Score Entropy (SE): Uncertainty sampling* is a common approach in active learning, which greedily selects queries involving (unobserved) user-item pairs for which the model is most uncertain (Settles, 2009). In our context, this corresponds to selecting the user-item pair with maximum *score entropy* w.r.t. the score distribution produced by the learned model. The rationale is clear: uncertainty in score predictions may lead to poor estimates of match quality. Of course, this approach fails to explicitly account for the matching objective (the term $Y(S^u, S^o)$ in Eq. 5), instead focusing (myopically) on entropy reduction in the predictive model (the term $\Pr(S^u|S^o, X, \theta)$).

*Score Max (SM):* An alternative, simple strategy is to select queries involving user-item pairs with highest predicted score w.r.t. MAP score estimates given our predictions of unobserved scores: $\hat{S}^u \equiv \arg\max_{S^u} \Pr(S^u|S^o, X, \theta)$. This may be especially advantageous for matching problems where scores for matched user-item pairs contribute an amount equal to their value in the matching objective (see Eq. 1).

An obvious shortcoming of both $S$E and $S$M is their insensitivity to the matching objective. Queries that reduce prediction entropy may have no influence on the resulting matching (e.g., if $s^u_{rp}$ has high entropy, but a much lower mean than some "competing" $s^u_{r'p}$, user $r'$ may remain matched to $p$ with high probability regardless of the response to query $rp$). One remedy is to use *expected value of information (EVOI)* to measure the improvement in matching quality given the response to a query (taking expectation over predicted responses). This approach has been used effectively in (non-constrained) CF (Boutilier et al., 2003); but EVOI is notoriously hard to evaluate. In our context, we would (in principle) have to consider each possible query $rp$, estimate the impact of each possible response $s^o_{rp}$ on the learned model (the term $\Pr(S^u|S^o, X, \theta)$ in

[1]Variances are sampled uniformly at random; in a real problem they would be given by the prediction model.

[2]In our settings one can elicit a rating or suitability score from a user for any item (e.g., paper, date, joke); so the full set $S^u_r$ serves as potential queries for user $r$.

Eq. 5), and re-solve the estimated matching (the term $Y(S^u, S^o)$ in Eq. 5). Instead, we consider several more tractable strategies.

*Y-Max (YM):* A simple way to select queries in a match-sensitive fashion is to consider the solution returned by the LP w.r.t. the observed scores, $S^o$, and the MAP solution of the unobserved scores, $\hat{S}^u$. We query the unknown pair $rp$ that contributes the most to the value of the objective: $\arg\max_{(rp)\in S^u} y_{rp}\hat{s}_{rp}$, where $y_{rp} \in Y(S^o, \hat{S}^u)$ is the binary match value for user r and item p, and $s_{rp}$ the corresponding MAP score value. In other words, we query the unobserved pair *among those actually matched* with the highest predicted score. We refer to this strategy as *Y-Max (YM)*. It reflects the intuition that we should either confirm or refute scores for matched pairs, i.e., those pairs that, under the current model, that directly determine the value of the matching objective. However, $Y$M is insensitive to score uncertainty.

*$\overline{Y}$-Max ($\overline{Y}$M)):* This method exploits our probabilistic matching model to select queries. As with $Y$M, $\overline{Y}$M queries the unobserved pair $rp$ that contributes the most to the objective value: $\arg\max_{(rp)\in S^u} \overline{Y}_{rp}\hat{s}_{rp}$. The difference is that we use the probabilistic match, exploiting prediction uncertainty in query selection.

*$\overline{Y}$-Entropy ($\overline{Y}$E):* This method exploits the probabilistic match $\overline{Y}$ as well, but unlike $\overline{Y}$M, $\overline{Y}$E queries unknown pairs whose entropy in the match distribution is greatest. Specifically, we view each $Y_{rp}$ as a Bernoulli random variable with (estimated) success probability $\overline{Y}_{rp}$. We then query that pair with maximum match entropy: $\arg\max_{(rp)\in S^u} \big[ -\overline{Y}_{rp}\log \Pr(\overline{Y}_{rp}) - (1-\overline{Y}_{rp})\log\Pr(1-\overline{Y}_{rp})\big]$.

One important point to note is that the match-sensitive strategies, $Y$M, $\overline{Y}$M, $\overline{Y}$E, all attempt to query unobserved pairs that occur (possibly stochastically) in the optimal match. When the LP does not match on any unobserved pairs, a fall-back strategy is needed. All three strategies resort to random querying as a fall-back, selecting a random unobserved item score for any specific user as its query. For $\overline{Y}$M and $\overline{Y}$E, we refer to any query that corresponds to a user-item pair with less than a 1% chance of being matched as a "random" query.

## 5. Experiments

We test the active learning approaches described above on three data sets, each with very different characteristics. We begin with a brief description of the data sets and matching tasks, then describe our experimental setup, before proceeding to a discussion of our results.

**Data sets** We first describe our three data sets and define the corresponding matching tasks.

*Jokes data set:* The Jester data set (Goldberg et al., 2001) is a standard CF data set in which over 60,000 users have each rated a subset of 100 jokes on a scale of -10 to 10. It has a dense subset in which all users rate ten common jokes. Our experiments use a data set consisting of these ten jokes and 300 randomly selected users. We convert this to a matching problem by requiring the assignment of a single joke to each user (e.g., to be told at a convention or conference), and requiring that each joke be matched to between 25 and 35 users (to ensure jocular diversity at the convention). Fig. 3(a) provides a histogram of the suitabilities for the Jester sub-data set.

*Conference data set:* This data is derived from the NIPS 2010 conference. It contains suitability scores for 1251 paper submissions provided for 48 area chairs (henceforth, reviewers). Scores range from 0 ("paper lies outside my expertise") to 3 ("very qualified to review"). Suitabilities for a subset of papers were elicited in two rounds. In the first round scores were elicited for about 80 papers per reviewer, with queries selected using the $YM$ procedure described above (where the initial scores were estimated using a simple language model (Balog et al., 2006) using reviewers' published papers). In the second round, unobserved scores were estimated using both the language model and a restricted Boltzmann machine (RBM) trained on the first-round scores and paper word-frequency vectors. Each reviewer was queried about 143 papers on average (excluding one outlier), and each paper received an average of 3.3 suitability assessments (std. dev. 1.3). The mean suitability score was 1.14 (std. dev. 1.1); a histogram of scores is shown in Fig. 3(b). Each paper was then assigned to one reviewer, and each reviewer received 20–30 papers.

*Dating data set:* The third data set comes from an online dating website (see `http://www.occamslab.com/petricek/data/`). It contains over 17 million ratings from roughly 135,000 users of 168,000 items (other users). We use a denser subset of 32,000 ratings from 250 users (each with at least 59 ratings) over 250 items (other users); see Fig. 3(c). Since items are users with preferences over their matches, dating is generally treated as a two-sided problem. While two-sided matching can fit within our general framework, the focus of our current work is on one-sided matching. As such, we only consider user preferences for "items" and not vice versa. Each user is assigned 25–35 items (and vice versa since "items" are users).

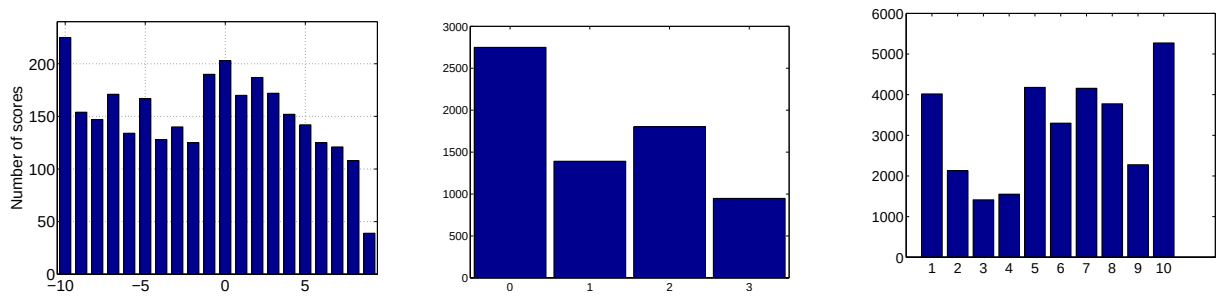

*Figure 3.* Histograms of known suitabilities for (a) Jokes, (b) Conference, and (c) Dating data sets.

### 5.1. Experimental Procedures

Our experiments simulate the typical interaction of a recommendation or matching engine with its users. All experiments start with a few observed preferences for each user and go through several rounds of querying. At each round, a querying strategy selects queries to ask one or more users. Note that in practice we restrict the strategies to only query (unobserved) scores available in our data sets. Once all users have responded, the system re-trains the learning model with newly and previously observed preferences, then proceeds to select the next batch of queries. This is a somewhat simplified model that assumes semi-synchronous user communication. We also assume for simplicity that the same fixed number of queries per user is asked in each round. The initial goal is simply to assess the relative performance of each method; we do relax some of these assumptions in Sec. 5.2.

There are a variety of reasonable interaction modes for eliciting user preferences. For example, in paper-reviewer matching, posing a single query per round is undesirable, since a reviewer, after assessing a single paper, must wait for other reviewer responses—and the system to re-train—before being asked a subsequent query. Reviewers generally prefer to assess their expertise off-line w.r.t. a *collection* of papers. Consequently batch interaction is most appropriate where, at each round, users are asked to assess $K$ items. While batch frameworks for active learning have received recent attention (e.g., (Guo & Schuurmans, 2007)), here we are interested in comparing different query strategies, hence use a very simple *greedy* batch approach where we elicit the "top" $K$ preferences from a user, where the "top" queries are ranked by the specific active strategy under consideration. Appropriate choice of $K$ is application dependent: smaller values of $K$ may lead to better recommendations with fewer queries, but require more frequent user interaction and user delay. We test different values of $K$ below.

We use BPMF to generate our predictions and its uncertainty model for unobserved scores. A procedure for setting some of the hyper-parameters of BPMF is outlined in (Salakhutdinov & Mnih, 2008a). We use a validation set for the other methods giving (using notation from the original paper) Jokes: $D = 1, \alpha = 0.1, \beta_{0u} = 0.1, \beta_{0v} = 10$; Conference: $D = 15, \alpha = 2, \beta_{0u} = \beta_{0v} = 0.1$; and Dating: $D = 2, \alpha = 2, \beta_{0u} = \beta_{0v} = 0.1$. Each observed score is assigned a fixed small uncertainty value of $1e^{-3}$. For $\overline{Y}$-based methods, which require sampling, we use 50 samples in all experiments.

We compare query selection methods w.r.t. their matching performance—i.e., the matching objective value of Eq. 1—using the match matrix given by the LP using estimated scores and known scores $S^o$, evaluated on the full set of available scores. We use a random querying strategy, which selects unobserved items uniformly at random for each user, as a *baseline*. All figures show the number of queries per user on the $x$-axis. The y-axis indicates the difference in the matching objective value between a specific querying strategy and the baseline. Positive differences indicate better performance relative to the baseline. The magnitude of this difference can be best understood relative to the number of users in the data set. For example, a difference of 300 in objective value for the 300 users in the Jokes data set means that users are better by one "score unit" on average. Note that as we increase the number of queries, even random queries will eventually find good matches—in the limit, where all scores are observed, matching performance of all methods will be identical (hence the bell-shape curves and asymptotic convergence in our results).

We don't focus on running time in our experiments since query determination can often be done off-line (depending on batch sizes). Having said that, even the most intense querying techniques are fast and can support online interaction: (a) in all 3 data sets, solving the LP takes a fraction of a second; (b) BPMF can be trained in a matter of a few minutes at most, but can be run asynchronously with query selection (which will use the most "up-to-date" learned model available); and (c) sampling scores is very fast as the posterior distribution is Gaussian. Furthermore, given the above, our methods should scale to larger datasets although the training time of BPMF may preclude fully online interaction.

### 5.2. Results

We first investigate the performance of the different querying strategies on our three data sets using default batch sizes—these $K$ values were deemed to be natural given the domains (different $K$ values are discussed below). Fig. 4(a) shows results for Jokes using batches of 10 queries per user per round ($K = 10$). Figs. 4(b) and (c) show Conference and Dating results, respectively, both with a batch size of 20. All users start with 20 observed scores: 15 are used for training and 5 for validation. We also experimented with a more realistic setting where some users have few observed scores (e.g., new users)—results are qualitatively very similar.

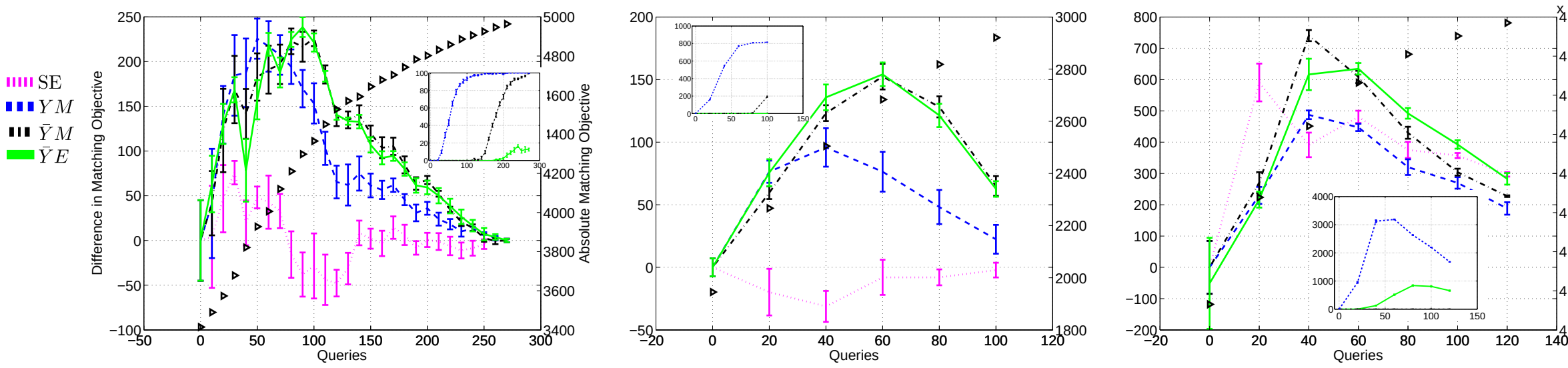

*Figure 4.* Matching performance for active learning results for (a) Jokes data set (10 queries per batch per user: qbu); (b) Conference (20 qbu); (c) Dating (20 qbu). Standard error is also shown. The overlaid figures represent the usage frequency of the fall-back query strategy for $Y$, $\overline{Y}$M and $\overline{Y}$E. The triangle plot (using the right vertical axis) shows the absolute matching value of the random strategy.

The relative performance of each of the active methods exhibits a fairly consistent pattern across all three domains, which permit us to draw some reasonably strong conclusions.[3] First, we see that all methods except for $S$E outperform the baseline in all domains. Recall that $S$E is essentially uncertainty sampling, a classic (match-insensitive) active learning model often used as a general baseline method for active learning. It outperforms the random baseline only occasionally, most significantly after the first round of elicitation in Dating. Second, all of our proposed match-sensitive techniques outperform $S$E consistently on all data sets. Third, the match-sensitive approaches that leverage uncertainty over scores, namely, $\overline{Y}$M and $\overline{Y}$E, typically outperform $Y$M, especially after the initial rounds of elicitation. This difference in performance behavior is most pronounced in the Conference domain.

We gain further insight into these results by examining the inner workings of these strategies. The overlay in Figs. 4(a–c) show the number of random (or fall-back) queries used (on average) by each of $Y$M, $\overline{Y}$M and $\overline{Y}$E. On all data sets $Y$M resorts to the fall-back strategy significantly earlier than the others, explaining $Y$M's fall-off in performance and indicating that the diversity of potential matches identified by our probabilistic matching technique plays a vital role in match-sensitive active learning.

Finally, when considering the performance of these methods on score prediction, we found no correlation between score prediction and matching performance. This further highlights the benefit of match-constrained active learning methods.

[3]We do not report the performance of $S$M— it is consistently outperformed by the baseline in all experiments. We have observed that $S$M typically selects all queries from among only a few items, namely, those with high predicted average score; hence it acquires no information about the vast majority of items.

**Sequential Querying** We employed a semi-synchronous querying procedure above, where all users are queried in parallel at each round. We now consider a different mode of interaction where, at each round, users are queried sequentially in round robin fashion. This allows the responses of earlier users within a round to influence the queries asked to later users—potentially reducing the total number of queries at the expense of increased synchronization (and delay) among users. Fig. 5(a) shows that our methods are robust to this modification in the querying procedure.

**Batch Sizes** The choice of the number of queries $K$ per batch affects both the frequency with which the user interacts with the system as well as the overall match performance. For example, high values of $K$ reduce the number of user "interactions" needed for a specific level of performance, at the expense of query efficiency (improvement in matching objective per query). The "optimal" value for $K$ depends on the actual recommendation application. Figs. 5(b) and (c) shows results with different values of $K$ on Conference, using 10 and 40 queries per round, respectively. The relative performance of the active methods remains almost identical. As expected, absolute performance w.r.t. query efficiency is better with smaller values of $K$. The matching-sensitive strategies clearly outperform the score-based techniques. Results are similar across all data sets.

**Matching constraints** Our results are also robust to different matching constraints, specifically, bounds on the numbers of items per user and vice versa (i.e., $R_{\min}, R_{\max}, P_{\min}, P_{\max}$). Using the Conference data set, we increase to two (from one) the number of reviewers assigned to each paper. Fig. 5(d) shows that the behavior of the methods changes little, with both $\overline{Y}$-methods still outperforming all other methods. The other domains (not shown) exhibit similar results.

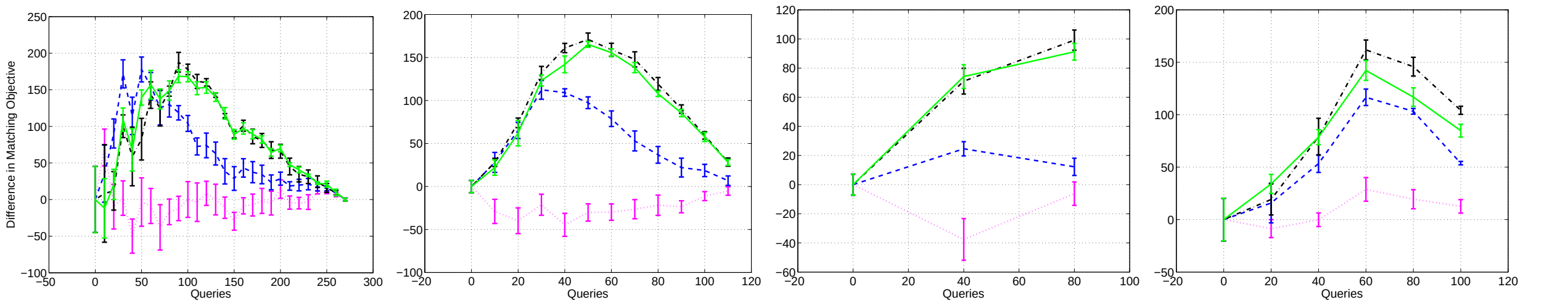

*Figure 5.* Matching performance for active learning results for: (a) Jokes data set (10 queries per batch per user, sequential); (b) Conference (10 queries per batch, parallel); (c) Conference (40 queries per batch, parallel); and (d) Conference data set with larger user and item constraints.

## 6. Conclusion

We investigated the problem of active learning for match-constrained recommendation systems. We explored several different approaches to generating queries that are guided by the matching objective, and introduced a novel method for probabilistic matching that accounts for uncertainty in predicted scores. Experiments demonstrate the effectiveness of our methods in determining high-quality matches with significantly less elicitation of user preferences than that required by uncertainty sampling, a standard active learning method. Our results highlight the importance of choosing queries in a manner that is sensitive to the matching objective and uncertainty over predicted scores.

There are many promising avenues of future research in match-constrained recommendation. We are currently exploring different matching objectives (e.g., two-sided matching with stability constraints) and methods for eliciting side-information from users in a way that is guided by the recommendation objective. Finally, higher-level, abstract queries (such as preferences over item categories or features) may significantly boost "gain per query" performance.

## References

Balog, K., Azzopardi, L., and de Rijke, M. Formal models for expert finding in enterprise corpora. *SIGIR-06*, pp.43–50, 2006.

Boutilier, C., Zemel, R. S., and Marlin, B. Active collaborative filtering. *UAI-03*, pp.98–106, 2003.

Brochu, E., Cora, V. M., and de Freitas, N. A tutorial on Bayesian optimization of expensive cost functions, with application to active user modeling and hierarchical reinforcement learning. TR-2009-23, CS, UBC, 2009.

Charlin, L., Zemel, R., and Boutilier, C. A framework for optimizing paper matching. *UAI-11*, pp.86–95, 2011.

Conry, D., Koren, Y., and Ramakrishnan, N. Recommender systems for the conference paper assignment problem. *RecSys'09*, pp.357–360, 2009.

Gale, D. and Shapley, L. S. College admissions and the stability of marriage. *Amer. Math. M.*, 69(1):9–15, 1962.

Goldberg, D., Nichols, D., Oki, B. M., and Terry, D. Using collaborative filtering to weave an information tapestry. *Comm. ACM*, 35:61–70, 1992.

Goldberg, K., Roeder, T., Gupta, D., and Perkins, C. Eigentaste: A Constant Time Collaborative Filtering Algorithm. *Inf. Retr*, 4:133–151, 2001.

Guo, Y. and Schuurmans, D. Discriminative batch mode active learning. *NIPS-07*, pp.593–600, 2007.

Harpale, A. and Yang, Y. Personalized active learning for collaborative filtering. *SIGIR-08*, pp.91–98, 2008.

Hylland, A. and Zeckhauser, R. J. The efficient allocation of individuals to positions. *Journal of Political Economy*, 87(2):293–314, 1979.

Jin, R. and Si, L. A Bayesian approach toward active learning for collaborative filtering. *UAI-04*, pp.278–285, 2004.

Koren, Y. The BellKor solution to the Netflix grand prize. *www.netflixprize.com/assets/GrandPrize2009_BPC_BellKor.pdf*, August, 2009.

McNee, S. M., Riedl, J., and Konstan, J. A. Being accurate is not enough: how accuracy metrics have hurt recommender systems. *CHI-06*, pp.1097–1101, 2006.

Mimno, D. M. and McCallum, A. Expertise modeling for matching papers with reviewers. *KDD-07*, pp.500–509, 2007.

Rigaux, P. An iterative rating method: application to web-based conference management. *SAC-04*, pp.1682–1687, 2004.

Rodriguez, M. A. and Bollen, J. An algorithm to determine peer-reviewers. *CIKM'08*, pp.319–328, 2008.

Roth, A. E. The evolution of the labor market for medical interns and residents: A case study in game theory. *J. Political Economy*, 92(6):991–1016, 1984.

Salakhutdinov, R. and Mnih, A. Bayesian probabilistic matrix factorization using Markov chain Monte Carlo. *ICML-08*, pp.880–887, 2008a.

Salakhutdinov, R. and Mnih, A. Probabilistic matrix factorization. *NIPS-08*, pp.1257–1264, 2008b.

Settles, B. Active learning literature survey. Computer Sciences TR-1648, Univ. Wisconsin-Madison, 2009.

Taylor, C. J. On the optimal assignment of conference papers to reviewers. TR MS-CIS-08-30, UPenn, 2008.

Weimer, M., Karatzoglou, A., Le, Q., and Smola, A. COFI RANK – Maximum margin matrix factorization for collaborative ranking. *NIPS-07*, pp.1593–1600, 2008.